\documentclass[letterpaper,10pt,preprint]{preprint}

\IEEEoverridecommandlockouts
\usepackage{graphicx}
\usepackage{algorithm}
\usepackage{algpseudocode}
\usepackage{amsmath}
\usepackage{amssymb}
\usepackage{url}

\title{\LARGE \bf
Imagine-RL: Residual-Confidence-Guided Cross-Attention for
World-Model-Augmented VLA Reinforcement Learning
}

\author{
Kejia Hu$^{1}$,
Wentong Zhai$^{2}$,
Bo Zhao$^{3}$,
and Shuai Liang$^{4,*}$
\thanks{$^{1}$Kejia Hu is with Department of Electric Engineering, Korea Advanced Institute of Science and Technology}
\thanks{$^{2}$Wentong Zhai is with University Of Science and Technology Beijing}
\thanks{$^{3}$Bo Zhao is with Shanghai Jiao Tong University}
\thanks{$^{4}$Shuai Liang is with Institute of Computing Technology, Chinese Academy of Sciences}
\thanks{$^{*}$Corresponding author: Shuai Liang
(\protect\url{liang.shuai.robotic@gmail.com}).}
}

\begin{document}

\maketitle
\thispagestyle{empty}
\pagestyle{empty}

\begin{abstract}

Reliable action evaluation in contact-rich manipulation requires looking beyond the current observation to future visual and contact consequences. Existing noise-space reinforcement learning efficiently steers a frozen Vision-Language-Action (VLA) policy, but its critics largely ignore these consequences. We present \textbf{Imagine-RL}, which augments noise-space VLA post-training with action-conditioned visual--torque imagination. For each candidate action chunk, a frozen visual--torque latent world model (\textbf{VTLWM}) autoregressively predicts compact future representations without pixel reconstruction. A current image--state--action query attends to observed histories and predicted futures, while previous-window prediction residuals provide token-wise confidence priors that suppress unreliable future tokens. By combining current evidence with predicted consequences, the action critic better evaluates candidate actions and supervises the actor, while the VLA and VTLWM remain frozen. Across four real-robot tasks with 50 evaluation trials per task, Imagine-RL uses only 100 RL trajectories and improves the average success rate by (23.6\%) over DSRL and by (60\%) over VLA baselines.

\end{abstract}

\section{Introduction}
Vision-Language-Action (VLA) models map language, images, and robot states to
actions and generalize across tasks and embodiments
\cite{black2024pi_0,kim2024openvla,team2024octo}. However, their predominantly
visual observations make contact-rich manipulation difficult. Force-aware
VLAs incorporate torque through multimodal fusion or auxiliary prediction
\cite{zhang2025ta,yu2026forcevla}, but only use it for action generation
rather than value estimation.

\begin{figure}[t]
  \centering
  \includegraphics[width=\linewidth]{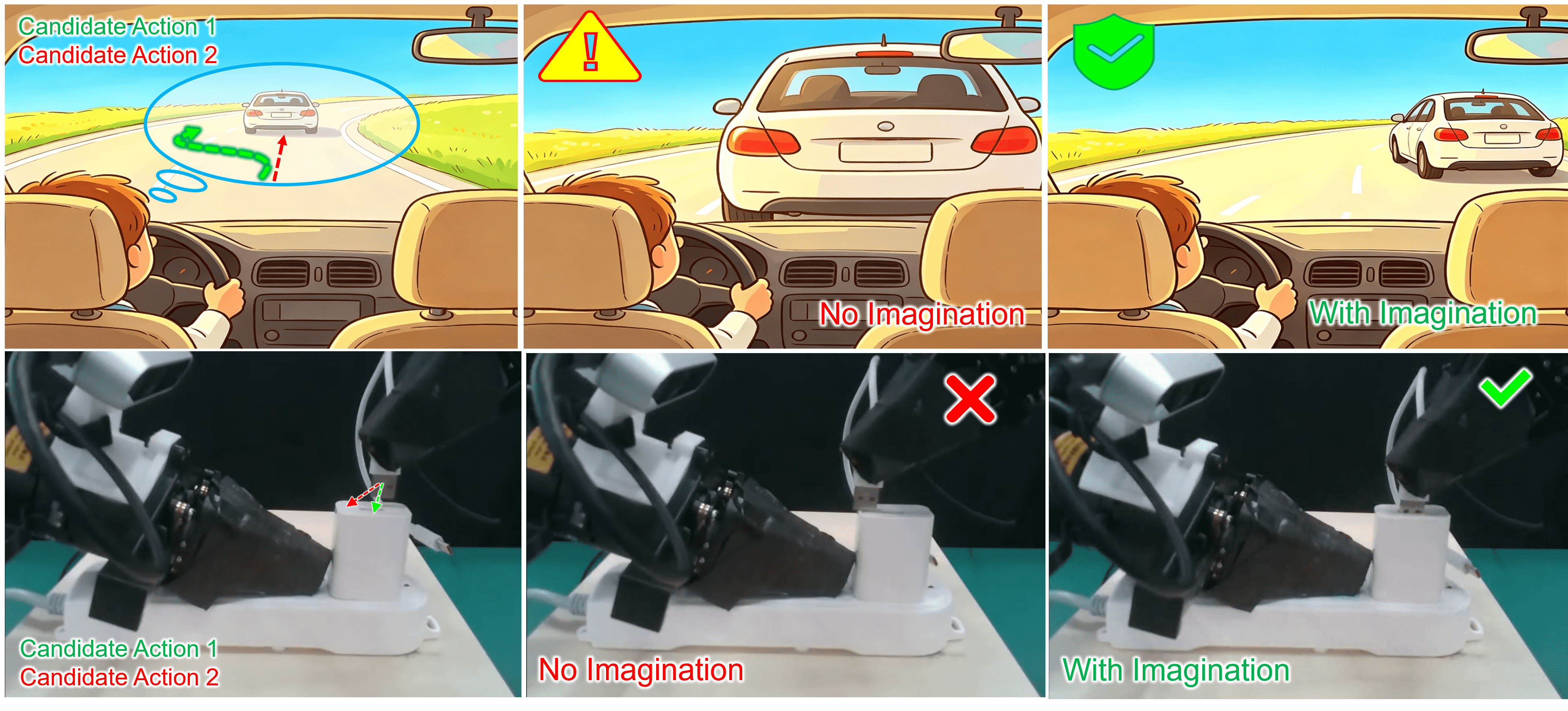}
  \caption{\textbf{Predictive critic evaluation.}
Both current and predicted future outcomes can be used to distinguish
candidate actions, as illustrated by the driving analogy. Without imagining
future consequences, two steering actions may appear similarly plausible.
By predicting their future trajectories, however, one can identify which
action leads to a safer outcome. Similarly, our critic evaluates candidate actions using both
the current reward and the predicted future visual--torque outcomes,
allowing it to distinguish actions that likely to succeed from those likely to fail.}
  \label{fig:teacher_critic}
\end{figure}

Reinforcement learning (RL) can adapt VLAs from task-level feedback
\cite{intelligence2025pi,li2025simplevla,ghasemipour2025self,
wagenmaker2025steering,xiao2025self,mark2024policy}, but real-robot interaction
is costly. Moreover, a conventional critic only scores according to current observation and candidate action. 
This is unreliable when visually similar actions are proposed, although they may lead to either success or failure. 
In latent-noise VLA-RL, the action critic also supervises the noise
critic and actor, making consequence-aware action ranking especially
important (Fig.~\ref{fig:teacher_critic}).

World models provide a natural source of such look-ahead evidence for reinforcement learning. Existing approaches use visual
futures, estimates values, or virtual interactions for policy improvement
\cite{wang2602gigabrain,zhu2025wmpo,xiao2025world,yang2026rise,xu2026rl}, among which
LeWorldModel predicts compact physical representations without heavy pixel
reconstruction \cite{maes2026leworldmodel}. Yet force-conditioned imagination
for critic learning remains underexplored.

In this paper, we propose \textbf{Imagine-RL}, which augments latent-noise VLA-RL with a
visual--torque latent world model (VTLWM). For each candidate action, VTLWM
predicts action-conditioned future representations. The critic attends to
visual and torque histories and futures together with current
image--state--action feature. Besides, residuals from the previous prediction window
become confidence priors on future-token attention, suppressing unreliable
imagination. The frozen VLA and VTLWM require neither backbone fine-tuning nor
pixel-level planning, leading to a light solution for contact-rich complicated tasks.
Real-robot experiments evaluate on task success rate, world-model prediction, critic
discrimination, and shared-candidate selection.

In this paper, we mainly make three contributions:
\begin{itemize}
\item  We develop Imagine-RL, which integrates action-conditioned visual--torque imagination into lightweight latent-noise VLA post-training.
\item We introduce VTLWM, a visual--torque latent world model that predicts action-conditioned future features for value estimation.
\item We design a residual-confidence-guided cross-attention \cite{lu2019vilbert} critic that down-weights unreliable future predictions while preserving observed history and current evidence.
\end{itemize}

\section{Related Works}
\subsection{VLA Foundations and Force-Aware VLAs}
Vision-language-action models transfer large-scale vision-language and
robot-data priors to general-purpose manipulation. Octo and OpenVLA
demonstrate broad cross-task transfer \cite{team2024octo,kim2024openvla},
while $\pi_0$ casts control as flow-based generation of continuous action
chunks \cite{black2024pi_0}. Yet these policies remain largely
vision-centric, leaving contact phenomena such as resistance, slippage,
and jamming only indirectly observable. ForceVLA addresses this gap with a
force-aware mixture-of-experts module \cite{yu2026forcevla}; TA-VLA instead
studies torque adapters and auxiliary torque prediction
\cite{zhang2025ta}. These approaches exploit force sensing as a useful
policy input. 

\subsection{RL Post-Training for VLAs}
Strong imitation priors do not eliminate failures caused by distribution
shift and compounding execution errors, motivating RL post-training from
task-level feedback. $\pi^*_{0.6}$ combines learned values with
advantage-conditioned policy learning \cite{intelligence2025pi}, whereas
SimpleVLA-RL studies scalable reinforcement learning for VLAs
\cite{li2025simplevla}. Other methods exploit autonomous experience or
residual-RL data generation \cite{ghasemipour2025self,xiao2025self}, and
policy-agnostic RL optimizes actions before distilling them into the policy
\cite{mark2024policy}. DSRL trains a lightweight actor that steers a frozen diffusion/flow policy only through its latent-noise space, instead of tuning the entire VLA \cite{wagenmaker2025steering}. 

\subsection{World Models for VLA-RL}
World models offer a complementary way to reduce physical trial-and-error
by anticipating action consequences. World-Env uses a learned environment
for VLA post-training \cite{xiao2025world}, while WMPO and GigaBrain-0.5M*
perform policy improvement with world-model-generated experience
\cite{zhu2025wmpo,wang2602gigabrain}. RISE separates action-conditioned
multi-view prediction from progress-value estimation and derives
advantages from imagined rollouts \cite{yang2026rise}. These methods show
the value of imagination, but generally rely on visually generated futures.
At a lighter level, RL Token demonstrates that compact representations can
connect large VLAs to online actor--critic learning \cite{xu2026rl}.
LeWorldModel learns action-conditioned latent dynamics from raw pixels using next-embedding prediction with Gaussian latent regularization, without for pixel reconstruction \cite{maes2026leworldmodel}. 
\section{Method}
\label{sec:method}
Imagine-RL is a lightweight reinforcement-learning framework that steers a frozen flow-matching VLA through its latent-noise space. A frozen visual--torque world model predicts the consequences of each decoded action chunk, and the resulting predictive values supervise a lightweight noise-space actor.

\begin{figure}[tbp]
    \centering
    \includegraphics[width=1.0\linewidth]{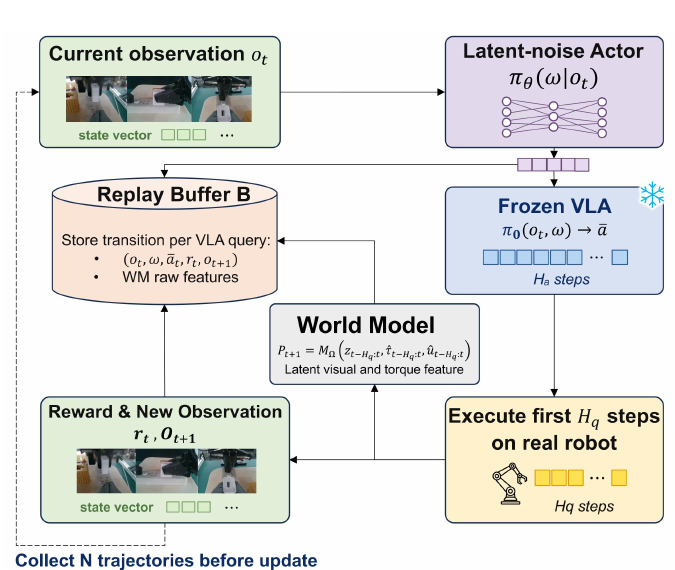}
    \caption{\textbf{Replay Buffer Collection of Imagine-RL}}
    \label{fig:data_collect}
\end{figure}
\begin{figure*}[t]
  \centering
  \includegraphics[width=0.98\textwidth]{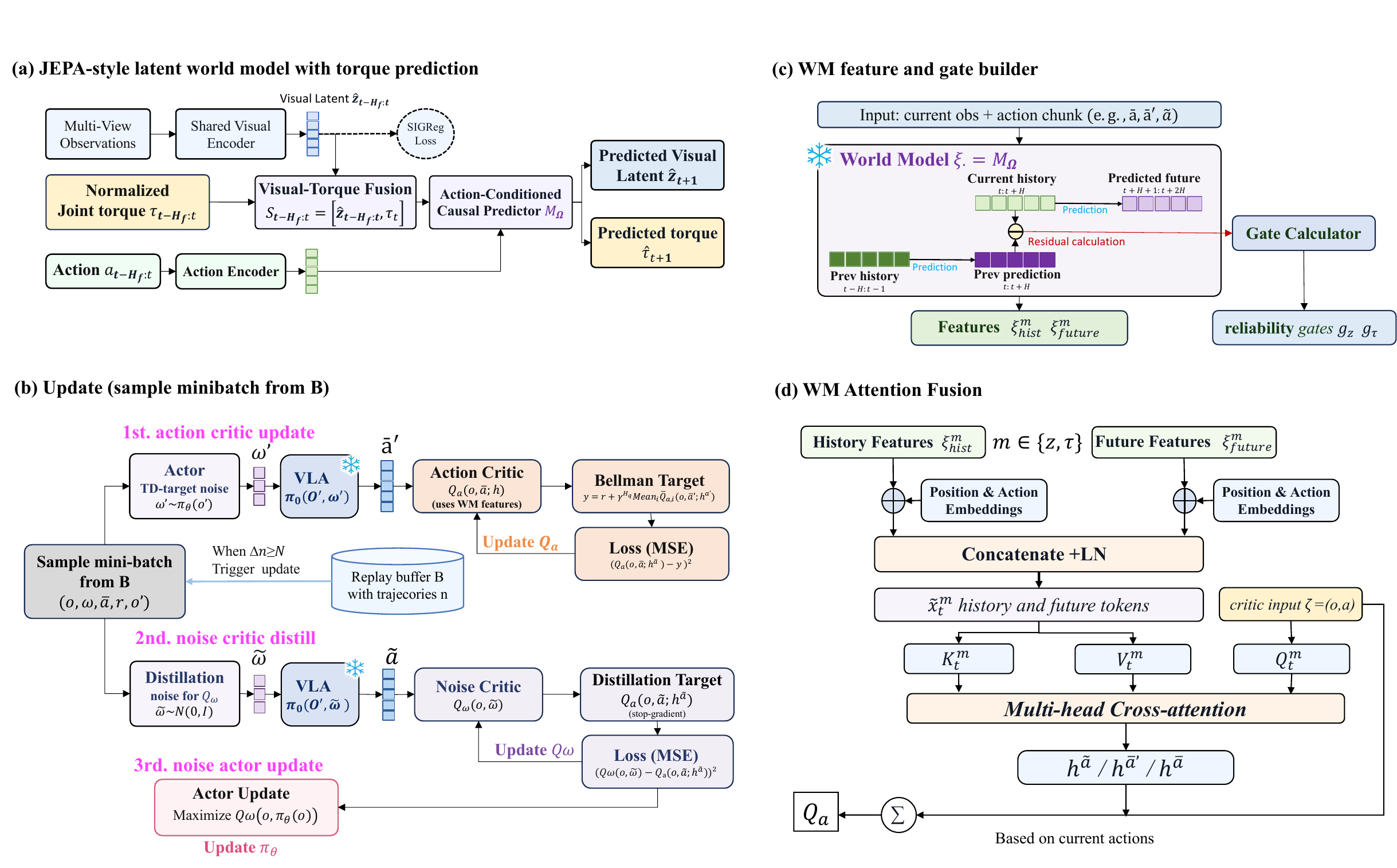}
  \caption{\textbf{Technical realization of the teacher-style predictive
  critic.} Read panels (a), (c), and (d) as one path: candidate action
   predicts visual--torque future and is used in 
  confidence-guided attention to give teacher action value. Panel (b)
  then shows how this value teaches the noise critic and actor. Thus, the
  action critic scores what the robot is doing now together with what that
  action is predicted to cause.}
  \label{fig:main_graph}
\end{figure*}

Fig.~\ref{fig:main_graph} presents the Imagine-RL framework. Panel (a) introduces VTLWM; panel (b) shows action-critic learning, noise-critic distillation, and noise-actor optimization; panel (c) constructs predicted features and residual-based prediction confidence; and panel (d) fuses observed and predicted visual--torque features for value estimation.

\begin{algorithm}[tbp]
\caption{Imagine-RL with Action-Aligned World-Model Using SAC \cite{haarnoja2017reinforcement,haarnoja2018soft,haarnoja2018soft2}}
\label{alg:main}
\footnotesize
\begin{algorithmic}
\State \textbf{Input:} frozen VLA policy $\pi_0$,
       frozen world model $M_\Omega$
\Statex \hspace{\algorithmicindent}
       offline data $\mathfrak{D}_{\mathrm{off}}$

\State Initialize replay buffer
       $\mathfrak{B}\gets\mathfrak{D}_{\mathrm{off}}$,
       action critic $Q_a$, target critic $\bar Q_a$,
       noise critic $Q_w$, and noise actor $\pi_\theta$

\State Initialize feature mapping $\Phi_\psi$ with frozen $M_\Omega$

\For{$t=1,\ldots,T$}

    \State Sample
    $(o_t,w_t,\bar a_t,r_t,o_{t+1})\sim\mathfrak{B}$
    \Statex \hspace{\algorithmicindent}
    where $\bar a_t=\pi_0(o_t,w_t)$ is the executed action
    
 \textbf{1st. Action-critic update}

    \State Sample
    $w'_{t+1}\sim\pi_\theta(o_{t+1})$
    and decode
    $\bar a'_{t+1}\gets\pi_0(o_{t+1},w'_{t+1})$

    \State Compute action-aligned critic features
    \Statex \hspace{\algorithmicindent}
    $h^{\bar a }_t\gets\Phi_\psi(o_t,\bar a_t)$,
    \quad
    $h^{\bar a'_{t+1}}_{t+1}\gets\Phi_\psi(o_{t+1},\bar a'_{t+1})$

    \State Update $Q_a$ by minimizing
    \Statex \hspace{\algorithmicindent}
    $\displaystyle
    \mathbb{E}\!\left[
      \left(
      Q_a(o_t,\bar a_t;h^{\bar a}_t)
      -r_t
      -\gamma^{H_q}
      \bar Q_a(o_{t+1},\bar a'_{t+1};h^{\bar a'_{t+1}}_{t+1})
      \right)^2
    \right]$

 \textbf{2nd. Noise-critic update}

    \State Sample
    $\tilde w_t\sim\mathcal{N}(0,I)$
    and decode
    $\tilde a_t\gets\pi_0(o_t,\tilde w_t)$
    \State Compute the candidate-aligned critic feature
    \Statex \hspace{\algorithmicindent}
    $h_t^{\tilde a_t}\gets\Phi_\psi(o_t,\tilde a_t)$

    \State Update $Q_w$ by minimizing
    \Statex \hspace{\algorithmicindent}
    $\displaystyle
    \mathbb{E}\!\left[
      \left(
      Q_w(o_t,\tilde w_t)
      -
      Q_a(o_t,\tilde a_t;h^{\tilde a_t}_t)
      \right)^2
    \right]$

 \textbf{3rd. Noise-actor update}

    \State Update $\pi_\theta$ by maximizing
    \Statex \hspace{\algorithmicindent}
    $\displaystyle
    \mathbb{E}_{o_t\sim\mathfrak{B}}
    \left[
      Q_w(o_t,\pi_\theta(o_t))
    \right]$

\EndFor

\end{algorithmic}
\end{algorithm}
\subsection{Latent-Noise VLA-RL}
\label{subsec:latent_noise_vla_rl}

Let $o_t$ denote the current visual observation at decision timestep $t$.
For brevity, robot-state conditioning is implicit in all policy and critic
expressions. The frozen VLA $\pi_0$ maps the observation and a latent-noise
input to a robot action chunk
\begin{equation}
    \bar{a}_t=[u_t,u_{t+1},\ldots,u_{t+H_a-1}],
    \qquad
    \bar{a}_t=\pi_0(o_t,w_t),
    \label{eq:vla_decode}
\end{equation}
where $H_a$ is the VLA output horizon. Imagine-RL does not learn robot actions directly. Instead, it learns a lightweight latent-noise actor
\begin{equation}
    w_t\sim\pi_\theta(w\mid o_t),
    \label{eq:latent_noise_policy}
\end{equation}
which steers the frozen VLA through its noise input.

We use two critics as shown in Fig.~\ref{fig:main_graph}(b). The action
critic $Q_a(o,\bar a;h^{\bar a})$ evaluates decoded robot-action chunks and
according to historical-current observations and world-model-generated future visual-torque feature $h^{\bar a}$, given $\bar{a}_t$. Then latent-noise critic $Q_w(o,w)$ is distilled from
$Q_a(o,\pi_0(o,w))$. This keeps the deployed policy lightweight: during RL, only noise actor $\pi_\theta$ is tuning and the base VLA is frozen.

The replay buffer, as shown in Fig.~\ref{fig:data_collect}, save one transition per VLA query, with chunk-level reward $r_t$ and discount $\gamma^{H_q}$. For $w_{t}\sim\pi_\theta(\cdot\mid o_{t})$ and $\bar a_{t}=\pi_0(o_{t},w_{t})$, we train multiple Q networks and get the mean in order to avoid bias
\begin{equation}
    y_t=
    r_t+
    \gamma^{H_q}\,
    \frac{1}{n}\sum_{i=1}^{n}
    \bar{Q}_{a,i}
    \left(
    o_{t+1},\bar a_{t+1};
    h_{t+1}^{\bar a'}
    \right),
    \label{eq:action_critic_target}
\end{equation}
where $n$ denotes the number of Q. 
The exact contents of $h_t^a$ are defined in Eq.\eqref{eq:attention_q_function}.

The noise critic is trained by querying the frozen VLA with sampled latent noise $\tilde w$:
\begin{equation}
\begin{aligned}
\min_{Q_w}\;
&
\mathbb{E}
\Bigl[
\Bigl(
Q_w(o_t,\tilde w_t)
-
Q_a
\bigl(
o_t,\tilde a_t;
 h^{\tilde a_t}_t
\bigr)
\Bigr)^2
\Bigr],
\\
&
\tilde a=\pi_0(o_t,\tilde w).
\end{aligned}
\label{eq:noise_critic_distill}
\end{equation}

The latent-noise actor is then updated to maximize $Q_w(o,\pi_\theta(o))$, with the usual entropy regularization in implementation. The complete training procedure is summarized in Algorithm~\ref{alg:main}.

\subsection{Force-Aware Latent World Model}
\label{subsec:force_aware_world_model}
To achieve better performance on contact-rich manipulation, Imagine-RL conditions the action critic on predictive features from a visual--torque latent world model (VTLWM). Extending LeWorldModel \cite{maes2026leworldmodel}, VTLWM jointly predicts visual and torque representations without pixel reconstruction: a shared encoder fuses multi-view images into $z_t$, while torque and action inputs are normalized as $\hat{\tau}_t$ and $ a_t$. We define
\begin{equation}
x_t=[z_t,\hat\tau_t],
\qquad e_t^a=E_a( a_t).
\label{eq:wm_state}
\end{equation}

We use a fixed one-step model with history length $H=15$. Given $H+1$
consecutive samples, the causal predictor uses the first $H$ state--action
pairs to predict $H+1$ state:

\begin{equation}
\begin{aligned}
P_{H}
&=M_\Omega\!\left(
X_{0:H-1},
{E_a}_{0:H-1}
\right), where\\
X_{0:H-1}
&=[x_0,\ldots,x_{H-1}], {E_a}_{0:H-1}=[{E_a}_0,\ldots,{E_a}_{H-1}]
\end{aligned}
\label{eq:wm_prediction}
\end{equation}

Here,
$M_\Omega$ is our world model, VTLWM. During rollout, the final prediction $P_H$ is
appended to the context and the inference window is shifted by one step, yielding an
autoregressive future sequence. VTLWM training loss is:
\begin{equation}
\begin{aligned}
\mathcal{L}_{\mathrm{WM}}
&=\operatorname{MSE}(P^z_{1:H},Z_{1:H})+\lambda_\tau
\operatorname{SmoothL1}(P^\tau_{1:H},T_{1:H})\\
&\quad
+\lambda_{\mathrm{sig}}\mathcal{L}_{\mathrm{SIGReg}}
\end{aligned}
\label{eq:wm_loss}
\end{equation}
where $Z_{1:H}$ and $T_{1:H}$ are the encoded visual and normalized force
ground-truth features.

During RL training, VTLWM is frozen and only generates critic features.
For each modality $m\in\{z,\tau\}$, the feature builder returns: (1) the current
history $X_t^{m,\mathrm{hist}}$, (2) a future prediction
$X_t^{m,\mathrm{fut}}(a_t)$ conditioned on the evaluated action, (3) a
previous-window prediction $P_t^{m,\mathrm{prev}}$.
\begin{equation}
\begin{aligned}
R_{t,j}^{m}
&=X_{t,j}^{m,\mathrm{hist}}
-P_{t,j}^{m,\mathrm{prev}},\\
\mathcal{S}_t^m(a_t)
&=\bigl\{X_t^{m,\mathrm{hist}},
X_t^{m,\mathrm{fut}}(a_t),\\
&\qquad P_t^{m,\mathrm{prev}},R_t^m\bigr\}.
\end{aligned}
\label{eq:wm_feature_set}
\end{equation}
History state, future prediction, and history-state-prediction residual masks indicate which positions are temporally
aligned. $j \in [0:H_f-1]$ is the index within each prediction horizon. We use $H=15$ history positions and recursively apply
the one-step predictor for $H_f=10$ future positions. The remaining future
positions are padded and excluded by the future mask.

\subsection{Confidence-Guided Cross-Attention Critic}
\label{subsec:confidence_gated_critic}

In order to reasonably utilize information, we design a Residual-Confidence-Guided Cross-Attention \cite{tan2019lxmert,jaegle2021perceiver} Critic, that scores a candidate action by selecting relevant visual and torque
information from observed history and predicted futures. Attention determines
what to retrieve, according to the prediction-error confidence that indicating how much to trust the
future. The retrieved information augments extra feature besides the current observation and action
for value estimation. Here we describe one modal $m$ as example. 

\paragraph{Query and tokens}
Let $\zeta_t=(o_t,a_t)$ denote the current visual observation paired with
the candidate action chunk ($\bar a_t$ or $\tilde a_t$ above).
A learned query map $f_q$ converts $\zeta_t$ as a $d=256$-dimensional
vector $q_t$:
\begin{equation}
q_t=f_q(\zeta_t).
\label{eq:attention_query}
\end{equation}
The world-model future is generated under that same action.

In Fig.~\ref{fig:main_graph}(d), $\xi_{\mathrm{hist}}^m$ and
$\xi_{\mathrm{future}}^m$ denote projected $X_t^{m,\mathrm{hist}}$ and
$X_t^{m,\mathrm{fut}}(a_t)$; adding position and action embeddings produces
the corresponding attention tokens. $P_t^{m,\mathrm{prev}}$ is
position-aligned with $X_t^{m,\mathrm{hist}}$ and is used only for residual
confidence estimation. Here $D_z=192$ and $D_\tau=14$; stacking the $H$
history and $H$ future tokens gives $\widetilde X_t^m$, with only $H_f$ valid
future positions.

\paragraph{Prediction reliability}
The previous-window features from the current timestep $t$, the historically predicted feature $P_{t,j}^{m,\mathrm{prev}}$ and the observed
history $X_{t,j}^{m,\mathrm{hist}}$ , indexed by $j$, are at the same timestamp;
their difference is the residual $R_{t,j}^m$ in Eq.\eqref{eq:wm_feature_set},
representing the WM generation reliability that persists across nearby windows. With the mean squared feature
magnitude $E_m(x)=\|x\|_2^2/D_m$, the relative error $\rho_{t,j}^m$ and generation reliability confidence $\alpha_{t,j}^m$ are
\begin{equation}
\begin{aligned}
\rho_{t,j}^m
&=\frac{E_m(R_{t,j}^m)}
{\tfrac12[E_m(X_{t,j}^{m,\mathrm{hist}})
+E_m(P_{t,j}^{m,\mathrm{prev}})]+\epsilon},\\
\alpha_{t,j}^m
&=\max\!\left(\alpha_{\min},e^{-\beta\rho_{t,j}^m}\right).
\end{aligned}
\label{eq:residual_confidence}
\end{equation}
This means smaller relative errors yield higher confidence. Here $\epsilon=10^{-6}$
prevents division by zero, $\beta=1$ controls sensitivity, and
$\alpha_{\min}=0.05$ is the confidence floor. Unavailable residuals receive
$\alpha_{\min}$ and $\rho=4$ for the statistics below.

\paragraph{From token relevance to context}
Separate learned projections produce a modality-specific query
$\mathbf Q_t^m=q_tW_Q^m$, matching keys
$\mathbf K_t^m=\widetilde X_t^mW_K^m$, and content values
$\mathbf V_t^m=\widetilde X_t^mW_V^m$. Keys and values represent the same
tokens: keys determine relevance, while values supply the information
retrieved, Fig.~\ref{fig:main_graph}(d). $Q_a$ denotes action value. Each modality has its own projection parameters;
affine biases are omitted.

The projected width $d$ is split into four heads indexed by $\ell$, each
with dimension $d_{\ell}=64$. The bias vector $\mathbf B_t^m$ has one entry per
token: $0$ for valid history, $\log\alpha_{t,j}^m$ for valid future step $j$,
and $-\infty$ for invalid slots. Each head computes
\begin{equation}
\begin{aligned}
\mathbf A_{t,\ell}^m
&=\operatorname{softmax}\!\left(
\frac{\mathbf Q_{t,\ell}^m(\mathbf K_{t,\ell}^m)^\top}{\sqrt{d_h}}
+\mathbf B_t^m
\right),\\
c_{t,\ell}^m&=\mathbf A_{t,\ell}^m\mathbf V_{t,\ell}^m,\\
c_t^m&=\operatorname{LN}\!\left(
\operatorname{Concat}_{\ell=1}^{4}(c_{t,\ell}^m)W_O^m
\right).
\end{aligned}
\label{eq:wm_cross_attention}
\end{equation}
Query--key dot products measure relevance; softmax converts the biased
scores into token weights $\mathbf A_{t,\ell}^m$ summing to one.
Their weighted sum of token values is the single-head context
$c_{t,\ell}^m\in\mathbb R^{64}$.
Concatenating the four contexts and applying the learned output projection
$W_O^m$ and layer normalization ($\operatorname{LN}$) yields
$c_t^m\in\mathbb R^{256}$.
Adding $\log\alpha$ discounts a future token's unnormalized weight by
$\alpha$, combining relevance with reliability.
Entirely invalid sequences produce zero context.

\paragraph{From contexts to action value}
Contexts $c_t^z,c_t^\tau$ summarize retrieved information.
The four reliability statistics $g_t^m$ are the mean, minimum, and maximum
confidence and mean relative error over valid future slots, or zeros if
none are valid. The core feature fusion is
\begin{equation}
\begin{aligned}
h_t^a&=[c_t^z,c_t^\tau,g_t^z,g_t^\tau],\\
Q_a(o_t,a_t;h_t^a)&=\mathcal Q([\zeta_t,h_t^a]).
\end{aligned}
\label{eq:attention_q_function}
\end{equation}
$\mathcal Q$ is a MLP critic outputting the Q value for policy optimization;
brackets concatenate the corresponding feature representations.
Thus, $\zeta_t=(o_t,a_t)$ supplies the current observation and action, while
$h_t^a$ adds the retrieved world-model contexts and their reliability.
Algorithm~\ref{alg:main} denotes this feature construction by $\Phi_\psi$,
with learned critic-side parameters $\psi$ and implicit history inputs.
We design multiple critics for stability and each learns its own updating; VTLWM remains frozen.

\section{Experiments}
\label{sec:experiments}

We conduct real-robot experiments to answer three questions:
(1) whether Imagine-RL improves task success rate over pretrained and
RL-post-trained VLA baselines;
(2) whether VTLWM predicts contact-relevant visual and torque dynamics;
and (3) whether residual-confidence-guided visual--torque attention improves
critic discrimination.
\begin{figure*}[t]
    \centering
\includegraphics[width=0.98\textwidth]{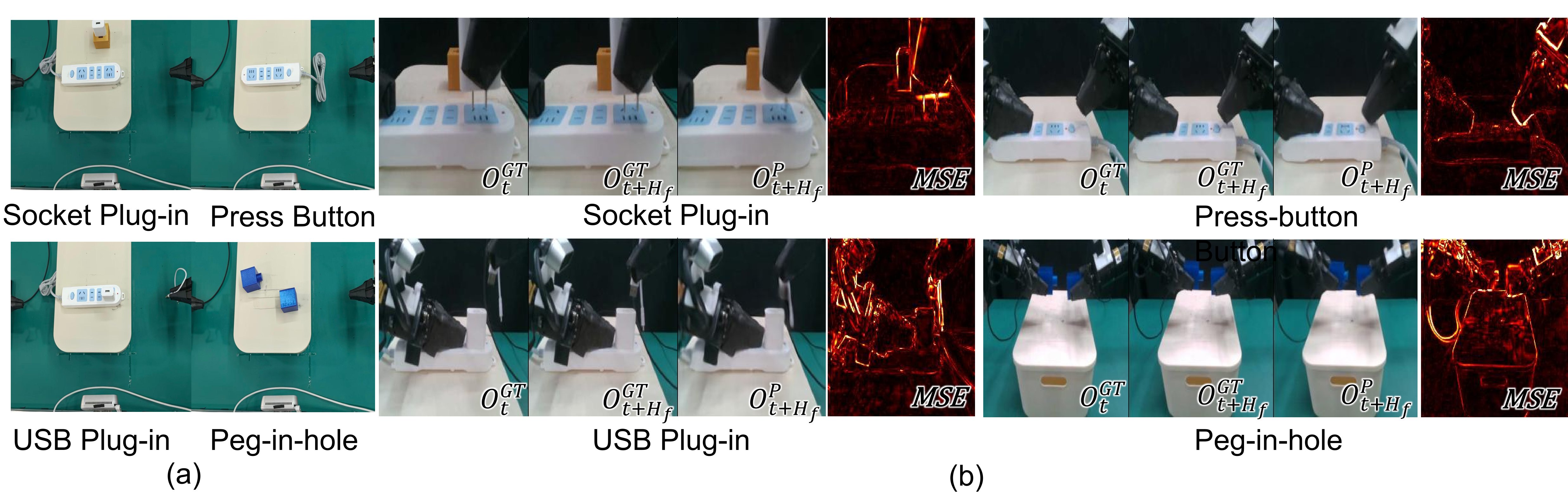}
    \caption{
    \textbf{Real-robot settings and qualitative results of action-conditioned visual prediction from our Latent World Model (key frames).}
    (a) Four contact-rich settings (Socket Plug-in, Press-button, USB Plug-in, Peg-in-hole).
    (b) VTLWM autoregressively predicted future representations $O^{P}_{t+H_f}$ and their corresponding ground-truth observations $O^{GT}_{t+H_f}$, conditioned with initial observation $O^{GT}_{t}$ and action $A_{t:t+(H_f-1)}$, here $H_f=10$. The MSE between $O^{GT}_{t+H_f}$ and $O^{P}_{t+H_f}$ are calculated and shown at end of each task.
    }
    \label{fig:settings_and_prediction}
\end{figure*}

\subsection{Experimental Setup / Results}
\label{subsec:experimental_setup}

\begin{figure}[tbp]
    \centering
    \includegraphics[width=\linewidth]
{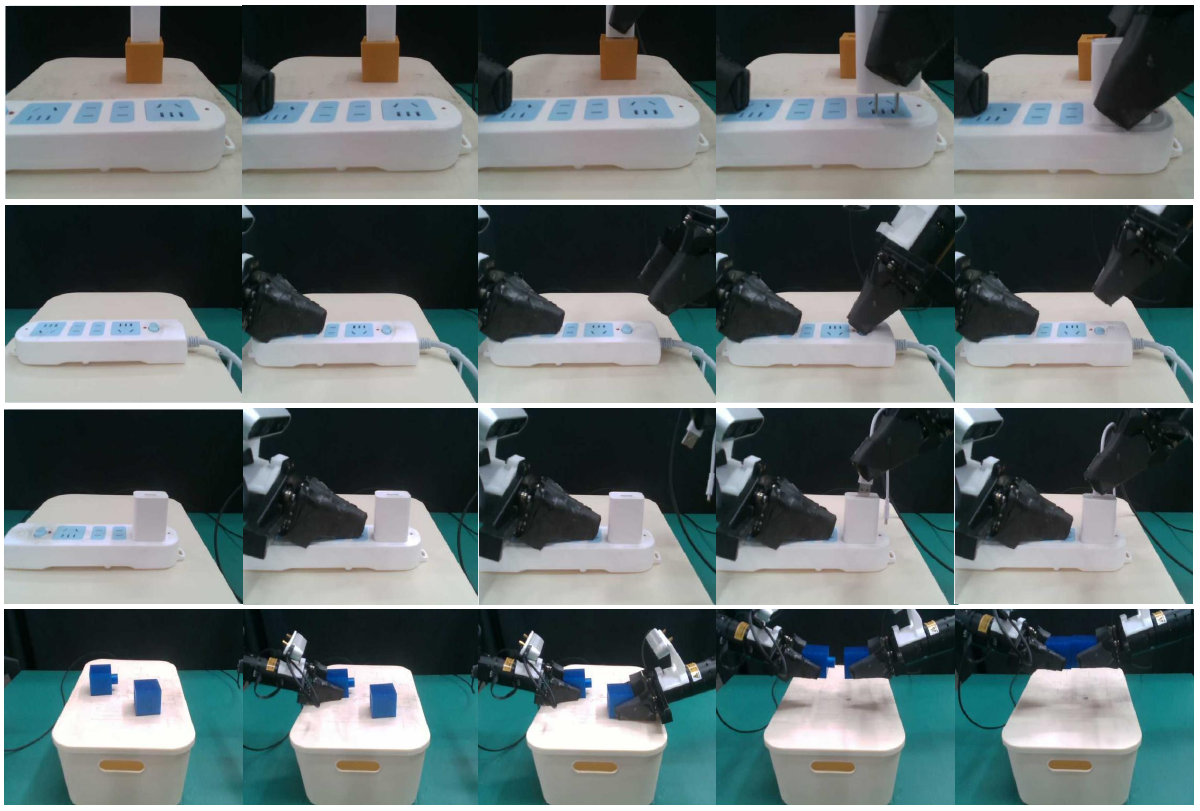}
    \caption{
    \textbf{Qualitative experiment results of four real robot tasks (Socket Plug-in, Press-button, USB Plug-in, Peg-in-hole).}
 Each task is a dual-arm collaborative manipulation task, including multiple stages within a single task, such as pressing, picking up, pressing again, and pushing.}
    \label{fig:four_task}
\end{figure}

\paragraph{Real-Robot Tasks.}
We evaluate all methods on four contact-rich manipulation tasks,
illustrated in Fig.~\ref{fig:settings_and_prediction}(a).
The tasks include
Socket Plug-in, Press-button,
USB Plug-in, and Peg-in-hole, Fig.~\ref{fig:four_task}. Each task is dual arm collaborative manipulation task, including multiple stage within one task, including pressing, picking up, pressing, pushing, etc.
For each task, we randomize object position and orientation within the same range
during evaluation.

\paragraph{Baselines.} We compare the following four methods:
\begin{itemize}
    \item \textbf{$\pi_0$}: the pretrained visual VLA without
    task-specific reinforcement learning;
    \item \textbf{TA-VLA}: a torque-aware VLA that incorporates force
    physical feedback/prediction;
    \item \textbf{DSRL}: latent-noise reinforcement learning with
    a world-model-free critic;
    \item \textbf{Imagine-RL}: our method, which augments latent-noise
    RL with residual-confidence-guided cross-attention over visual and
    torque histories and candidate-action-conditioned futures through our Latent World Model.
\end{itemize}

\paragraph{Training Protocol.}
For each task, the base $\pi_0$ is pre-trained with $100$ trajectories.
Imagine-RL and DSRL use the same pretrained VLA backbone and an initial
warm-up buffer of $50$ trajectories. Both methods receive the same amount of
online interaction. At each non-terminal query transition, the reward is
$-1$; successful termination gives $+5$, ordinary failed truncation gives
$-1$, and a severe absorbing failure gives $-50$. We use two NVIDIA RTX 3090
GPUs to host the remote base VLA ($\pi_0$) server, while the RL components are
trained locally on a single NVIDIA RTX 3090 GPU.

\paragraph{Implementation Details.}
Our robotic platform is a PIPER robotic arm equipped with Intel RealSense D435i cameras.
The frozen VLA outputs an action chunk of length $H_a=30$, of which the first
$H_q=10$ actions are executed before replanning. VTLWM uses a history length
$H=15$ and a one-step predictor, recursively rolled out for $H_f=10$ valid
future visual and torque states at 30 Hz. The visual and torque token dimensions are 192 and
14, respectively. Cross-attention has width 256 and four heads, with no
attention dropout. The action critic uses an ensemble of ten Q-functions and
mean target reduction. We use a batch size of 64, a per-control-step discount
of 0.999, actor learning rate $5\times10^{-5}$, critic-encoder learning rate
$10^{-4}$, and critic-head/fusion learning rate $3\times10^{-4}$. RGB inputs
from the left and right wrist cameras and the top-view camera are resized to
$128\times128$.

\begin{figure}[tbp]
    \centering
    \includegraphics[width=\linewidth]
    {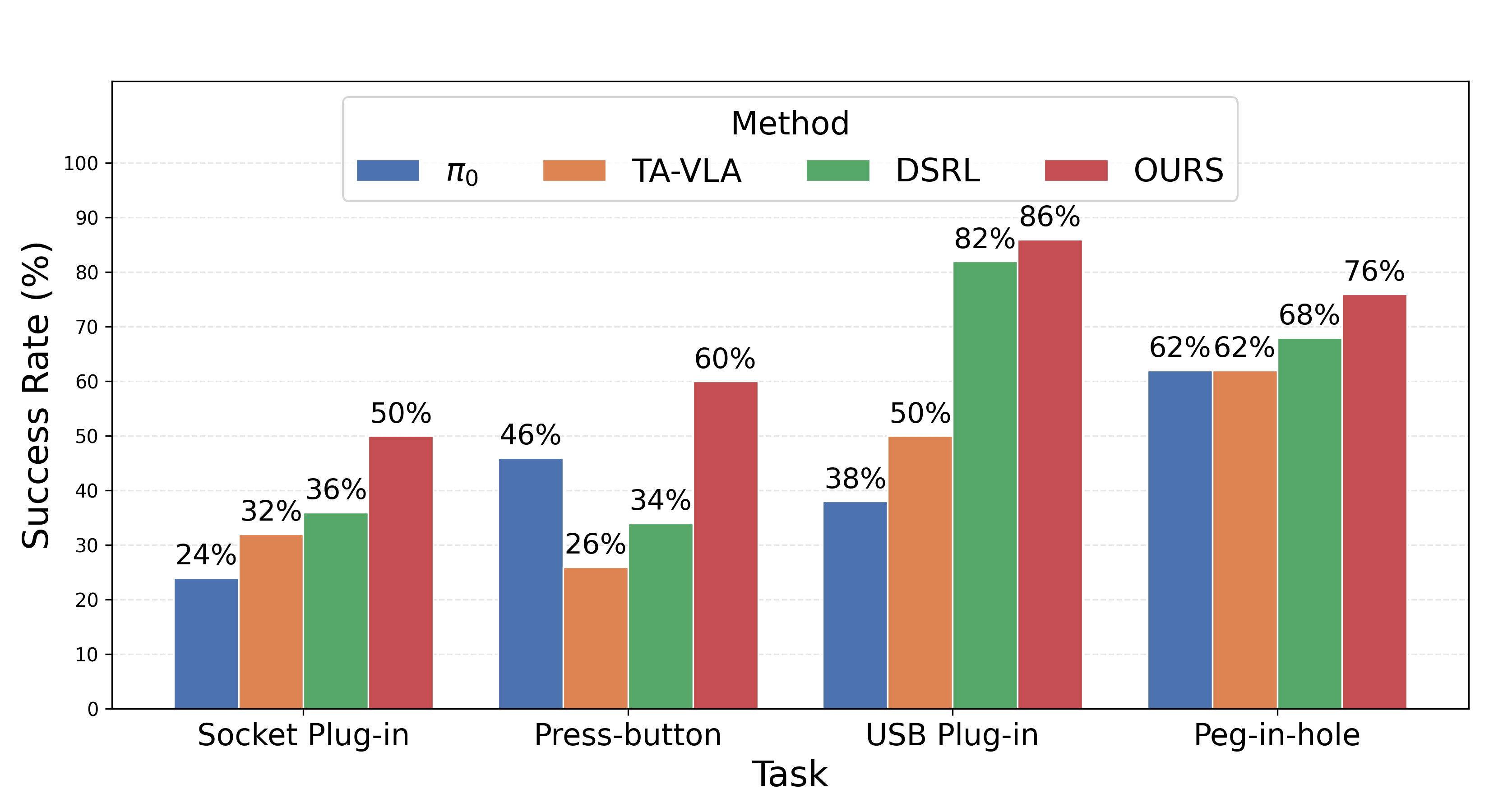}
    \caption{Success rates on the four real-robot tasks (Socket Plug-in, Press-button, USB Plug-in ,and
Peg-in-hole, each task 50 trials), 50 episodes RL training (every 10 episodes updating once), after 50 episodes warm-up RL training.}
    \label{fig:success_rate}
\end{figure}

Before the first RL update, we collect 50 pure-$\pi_0$ trajectories and run
at most 5000 initial gradient steps. Thereafter, RL training is performed after
every ten newly collected trajectories. All methods use identical RL
hyperparameters unless otherwise stated.
\paragraph{Results.}
We evaluate each method for $50$ trials per task, with the training procedure of every 10 episodes updating once. The initial object poses and task conditions are kept the same for all
methods. Success rates are shown in Fig.~\ref{fig:success_rate}.
For the two imitation methods, our method achieves an average relative success rate rise of $60\%$ compared to $\pi_0$ and $60\%$ compared to TA-VLA. For the reinforcement learning baseline DSRL, we achieve a rise of $23.6\%$, within only 100 episodes updating (50 episodes warm-up, $10*5$ episodes updates).

\begin{figure}[tbp]
    \centering
    \includegraphics[width=0.8\linewidth]{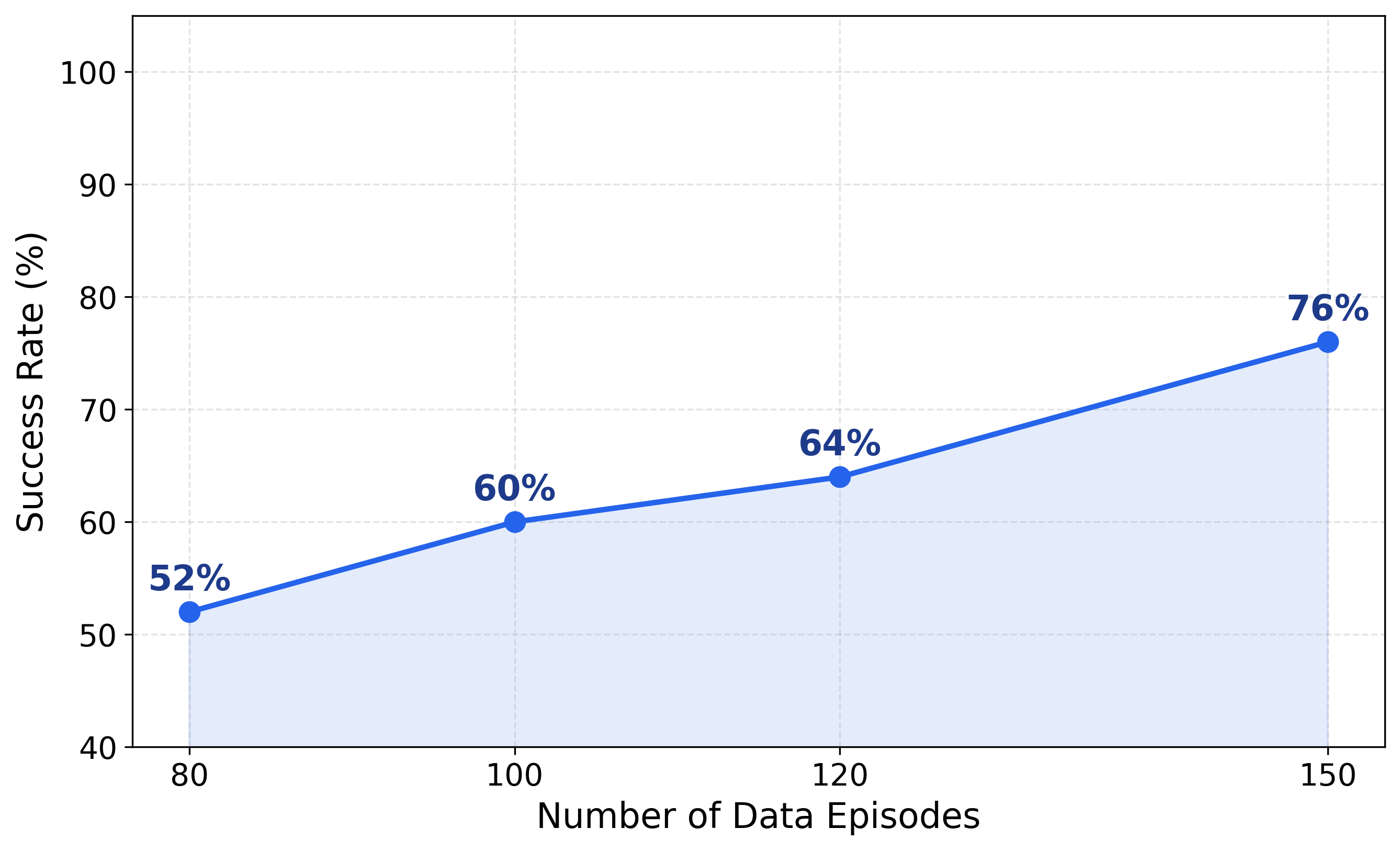}
    \caption{Success rate with increasing online interaction data.}
    \label{fig:success_rise}
\end{figure}

To examine performance with additional interaction, we continue training
Imagine-RL on the \textit{press-button} task until the online buffer up to
$100$ trajectories ($10*10$ episodes updating) after warm-up training. Fig.~\ref{fig:success_rise} shows the resulting success
rate as the amount of online data increases, which achieves $+46\%$ success rate increment for task \textbf{Press-button}, within only $10*10$ episodes RL updating after $50$ epidodes warm up training, showing the capability of our method can improve fast within only small numbers RL trials.

\subsection{World-Model Prediction Analysis}
\label{subsec:wm_analysis}

We next test whether VTLWM captures the key future dynamics required by
the critic. We evaluate the frozen world model on held-out-robot
sequences.

\paragraph{Visual Future Prediction.}
Fig.~\ref{fig:settings_and_prediction}(b) shows the qualitative visual feature prediction performance of VTLWM on four tasks (Socket Plug-in, Press-button, USB Plug-in, Peg-in-hole), comparing predicted pixel visual futures decoded by a trained decoder with ground-truth observations. The decoder is not used by the critic or the deployed policy. VTLWM autoregressively predicted future visual representation $O^{P}_{t+H_f}$ and their corresponding ground-truth observations $O^{GT}_{t+H_f}$, conditioned with initial visual observation $O^{GT}_{t}$, torque $T^{GT}_{t}$ and action $A_{t:t+(H_f-1)}$, here $H_f=10$. 
As shown in Fig.~\ref{fig:settings_and_prediction}(b), VTLWM captures gripper-object interaction showing consistency with ground-truth as the action proceeding within prediction horizon.
Although prediction quality gradually decreases with rollout length resulting from the compounding error through VTLWM autoregressive inference, the predicted representations still preserve the task-relevant temporal and spatial features between the robot and the environment as action evolves, which is determinative information for critic estimation.


\paragraph{Torque Future Prediction}

\begin{figure}[tbp]
    \centering
    \includegraphics[width=\linewidth]
{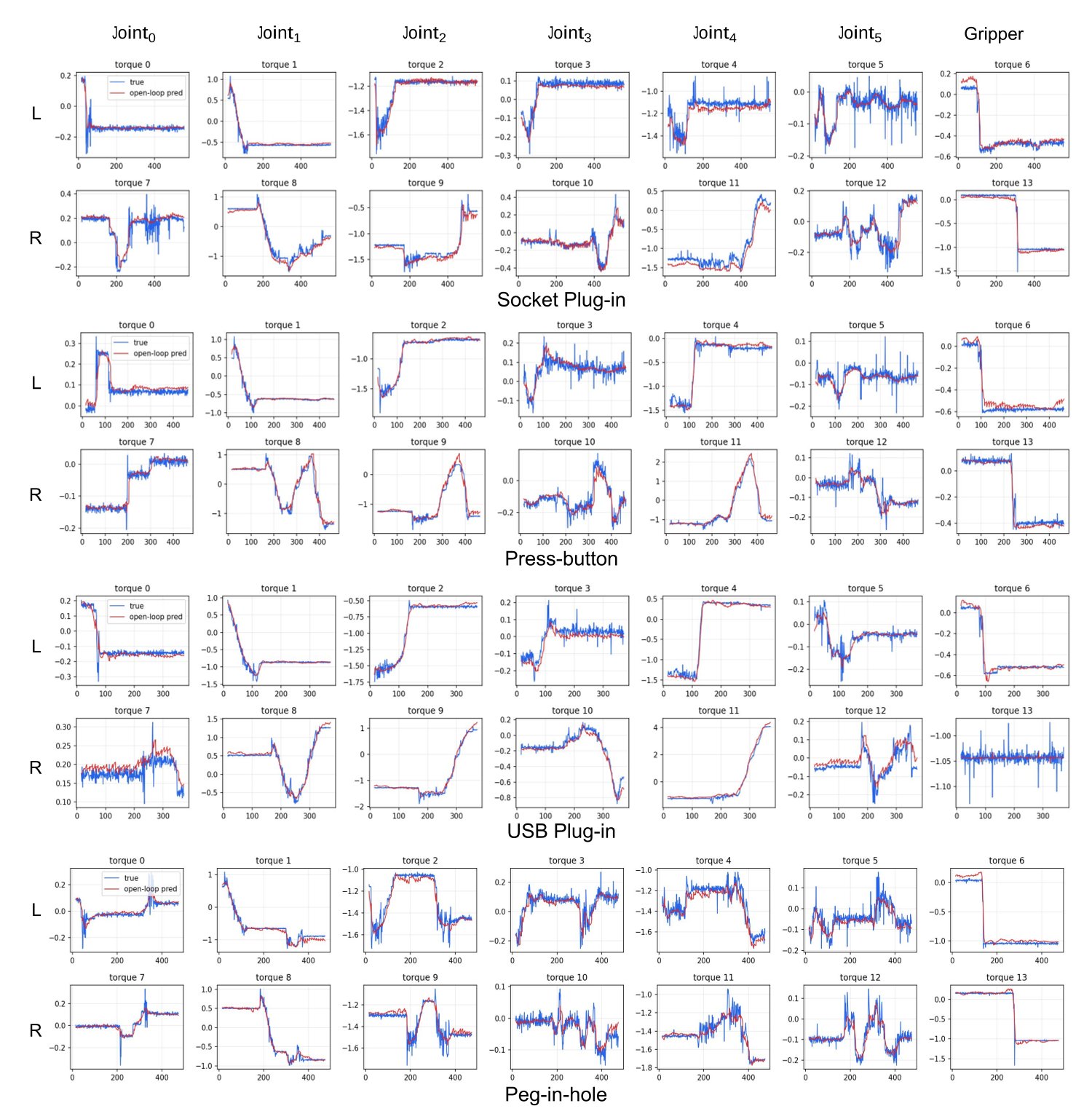}
    \caption{
    \textbf{Torque prediction on held-out real-robot trajectories on four tasks (Socket Plug-in, Press-button, USB Plug-in, Peg-in-hole).}
    Red lines denote ground-truth measured torque and blue lines denote VTLWM predictions. $L$ and $R$ represents the left and right robot arms.
    }
    \label{fig:torque_prediction}
\end{figure}

Fig.~\ref{fig:torque_prediction} shows the qualitative torque feature prediction performance of VTLWM on four tasks (Socket Plug-in, Press-button, USB Plug-in, Peg-in-hole), compared with ground-truth torque values on 12 (6*2) joints and 2 grippers. 
\begin{figure*}[htbp]
    \centering
    \includegraphics[width=0.77\textwidth]{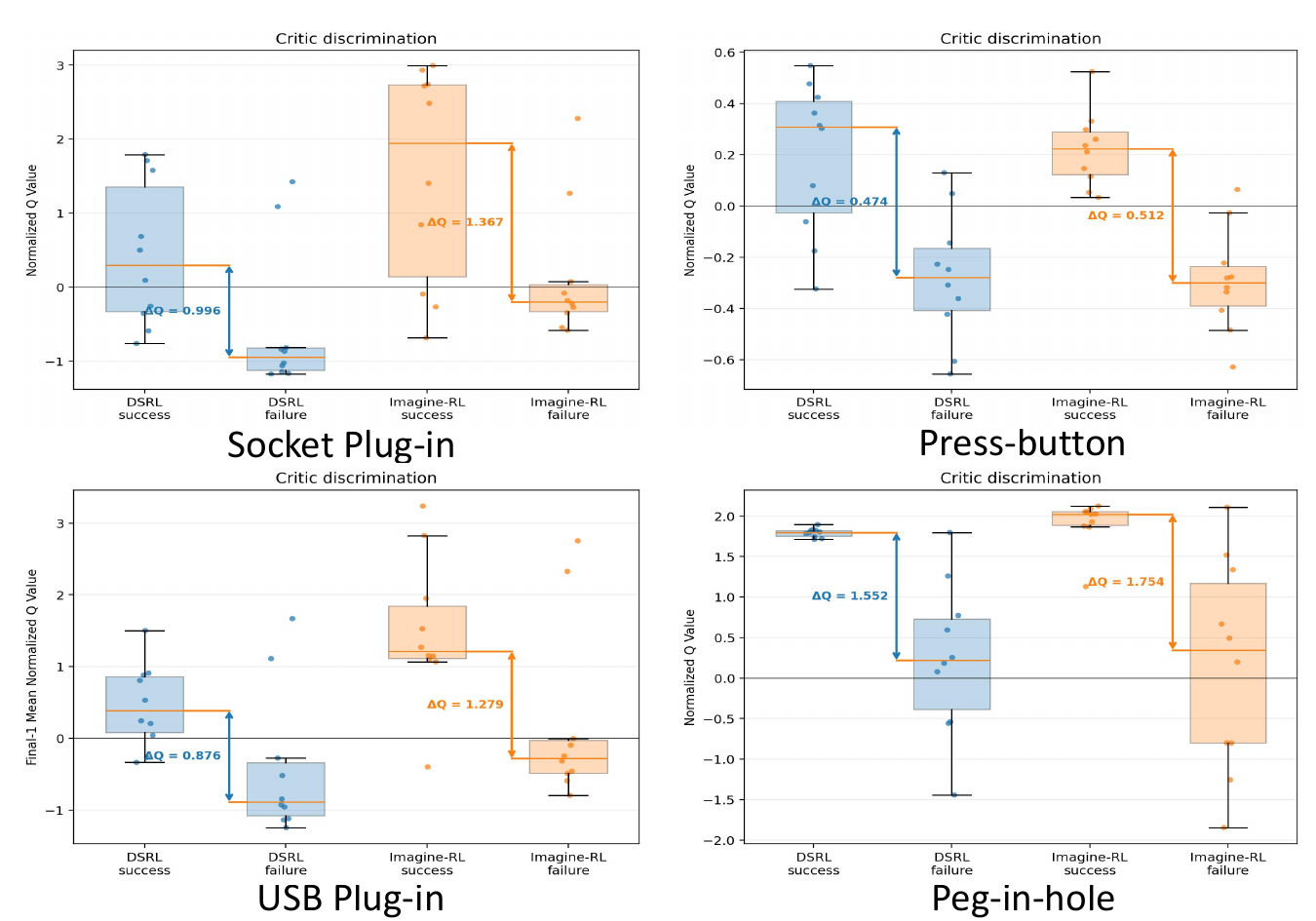}
    \caption{
    \textbf{Critic discrimination analysis.}
    Results on Socket Plug-in, Press-button, USB Plug-in, and Peg-in-hole. For
    each task, DSRL and Imagine-RL score the same executed action chunks from
    10 held-out successful and 10 failed trajectories under the same reward
    setting. Q-values are z-score standardized separately for each critic within each
    task, and boxes summarize the trajectory-level distributions. For all four
    tasks, each dot is the standardized Q-value at the final retained VLA query
    step of one trajectory. The annotated margin is
    $\Delta Q=\mathbb{E}[\bar{Q}\mid\mathrm{success}]
    -\mathbb{E}[\bar{Q}\mid\mathrm{failure}]$. The DSRL and Imagine-RL
    margins are, respectively, $0.996$ and $1.367$ on Socket Plug-in,
    $0.474$ and $0.512$ on Press-button, $0.876$ and $1.279$ on USB Plug-in,
    and $1.552$ and $1.754$ on Peg-in-hole. These correspond to relative
    increases of $37.2\%$, $8.0\%$, $46.0\%$, and $13.0\%$, respectively,
    demonstrating consistently stronger critic discrimination by Imagine-RL.
    }
    \label{fig:critics_analysis}
\end{figure*}
\begin{table}[htbp]
\caption{Cosine similarity of visual predictions}
\label{tab:cosine_value}
\begin{center}
\begin{tabular}{|c||c|c|c|c|c|}
\hline
Method & Socket & Press & USB & Peg & Avg. \\
\hline
Visual & 0.955 & 0.915 & 0.881 & 0.901 & 0.913 \\
\hline
\end{tabular}
\end{center}
\end{table}

\begin{table}[htbp]
\caption{Torque prediction performance on MAE, MSE, RMSE of four real robot task}
\label{tab:force_statics}
\begin{center}
\begin{tabular}{|c||c|c|c|c|c|}
\hline
Method & Socket & Press & USB & Peg & Avg. \\
\hline
MAE & 0.0483 & 0.0455 & 0.0449 & 0.0378 & 0.0441 \\
\hline
MSE & 0.0059 & 0.0070 & 0.0053 & 0.0035 & 0.0054 \\
\hline
RMSE & 0.0770 & 0.0834 & 0.0731 & 0.0589 & 0.073 \\
\hline
\end{tabular}
\end{center}
\end{table}VTLWM autoregressively predicted future torque $T^{P}_{t+H_f}$ and their corresponding ground-truth torque $T^{GT}_{t+H_f}$, conditioned with initial visual observation $O^{GT}_{t}$, torque $T^{GT}_{t}$ and action $A_{t:t+(H_f-1)}$, here $H_f=10$. 
The predicted force signals are able to effectively capture the onset and temporal evolution of contact events and key robot motion, providing extra action trigger signal for critic evaluation, and complementing information that is only weakly observable from RGB inputs. 
For example, during the press-button task, whether a press is effective can be determined by sensing the vertical elastic force, which continuously increases with the pressing depth and suddenly disappears when the button locks. Such contact dynamics are difficult to capture using visual sensing alone.

We further evaluate VTLWM quantitatively. All the data are from the 50 warm-up and 10*5 RL trials from the frozen pre-trained VTLWM. 
We calculate the mean visual cosine similarities of the latent embedding between the prediction and ground-truth across the whole four tasks, as reported in Table~\ref{tab:cosine_value}. From Table~\ref{tab:cosine_value} we can find that the visual cosine similarities between ground truth and VTLWM predicted embedding are very close to 1, illustrating that our VTLWM is able to capture the dynamics and predict the reasonable future evolution in advance, providing key information for critic.
Furthermore, we calculate the MAE, MSE, RMSE of torque prediction on each task, as shown in Table~\ref{tab:force_statics}. VTLWM achieves relatively high predictive accuracy, with an MAE of 0.0441, corresponding to only $1.00\%$ of the maximum value (±4.3799).


\subsection{Critic Discrimination}
\label{subsec:critics_analysis}
We examine whether the action critic can distinguish executed actions from
successful and failed interactions under the same reward definition and
scale. For each of the four tasks, we use the same held-out set of 10
successful and 10 failed trajectories for DSRL and Imagine-RL. At every
retained VLA query step, both critics receive the same observation and score
the same executed action chunk of length $H_q$; Imagine-RL additionally
conditions its evaluation on the action-conditioned visual--torque features
predicted by VTLWM. Thus, the comparison isolates how well the learned value
functions discriminate action outcomes rather than differences in the
evaluated actions or rewards.

Because the two critics are trained independently and their raw Q-values need
not share a common scale, within each task we standardize each critic as
$\bar Q=(Q-\mu_Q)/(\sigma_Q+\epsilon)$, with $\mu_Q$ and $\sigma_Q$ computed
over all query-step Q-values from both outcome sets. Each point in
Fig.~\ref{fig:critics_analysis} is the standardized Q-value at the final
retained VLA query step of one trajectory. We use this same
terminal-interaction statistic for all four tasks and quantify discrimination
as
$\Delta Q=\mathbb{E}[\bar{Q}\mid\mathrm{success}]
-\mathbb{E}[\bar{Q}\mid\mathrm{failure}]$, where $\bar{Q}$ denotes the
within-critic standardized value. A larger $\Delta Q$ therefore indicates a
stronger separation between actions associated with successful and failed
outcomes. Imagine-RL consistently produces a larger margin than DSRL on all
four tasks. Averaged across the four tasks, the discrimination margin increases
from $\mathbf{0.975}$ for DSRL to $\mathbf{1.228}$ for Imagine-RL, corresponding
to a $\mathbf{26.0\%}$ relative improvement. These results show that the
VTLWM-conditioned critic assigns more distinctive values to successful and
failed interaction actions under an identical reward setting.

\section{Conclusion}

We presented Imagine-RL, a lightweight approach for adapting a frozen VLA to
contact-rich manipulation through latent-noise reinforcement learning. A
visual--torque latent world model (VTLWM) predicts the consequences of each action
chunk evaluated by the critic. The action critic uses the current
image--state--action representation to query visual and torque history/future
tokens, while verified past prediction residuals suppress unreliable future
tokens through an additive attention prior. This action-aligned predictive
critic provides the teacher values used to train the lightweight noise critic
and actor without fine-tuning the VLA backbone or performing pixel-level
planning. Our real-robot evaluations examine the resulting task performance, visual-torque latent world model prediction performance and critic discrimination. Experiment results show that our Imagine-RL method achieves faster convergence speed and higher success rate compared with baselines. Future work will investigate continual world-model adaptation under policy-induced distribution shift, multi-timescale prediction of delayed contact outcomes, and uncertainty-aware exploration with safe fallback to the frozen VLA prior.

\section*{Acknowledgment}
The driving illustration in Fig.~\ref{fig:teacher_critic} was generated using
Doubao AI. OpenAI Codex was used to assist with portions of the code
implementation. All AI-assisted content was reviewed and verified by the
authors.




\bibliographystyle{IEEEtran}
\bibliography{references}

\begin{thebibliography}{10}
\providecommand{\url}[1]{#1}
\csname url@samestyle\endcsname
\providecommand{\newblock}{\relax}
\providecommand{\bibinfo}[2]{#2}
\providecommand{\BIBentrySTDinterwordspacing}{\spaceskip=0pt\relax}
\providecommand{\BIBentryALTinterwordstretchfactor}{4}
\providecommand{\BIBentryALTinterwordspacing}{\spaceskip=\fontdimen2\font plus
\BIBentryALTinterwordstretchfactor\fontdimen3\font minus
  \fontdimen4\font\relax}
\providecommand{\BIBforeignlanguage}[2]{{%
\expandafter\ifx\csname l@#1\endcsname\relax
\typeout{** WARNING: IEEEtran.bst: No hyphenation pattern has been}%
\typeout{** loaded for the language `#1'. Using the pattern for}%
\typeout{** the default language instead.}%
\else
\language=\csname l@#1\endcsname
\fi
#2}}
\providecommand{\BIBdecl}{\relax}
\BIBdecl

\bibitem{black2024pi_0}
K.~Black, N.~Brown, D.~Driess, A.~Esmail, M.~Equi, C.~Finn, N.~Fusai, L.~Groom,
  K.~Hausman, B.~Ichter \emph{et~al.}, ``$pi_0$: A vision-language-action flow
  model for general robot control,'' \emph{arXiv preprint arXiv:2410.24164},
  2024.

\bibitem{kim2024openvla}
M.~J. Kim, K.~Pertsch, S.~Karamcheti, T.~Xiao, A.~Balakrishna, S.~Nair,
  R.~Rafailov, E.~Foster, G.~Lam, P.~Sanketi \emph{et~al.}, ``Openvla: An
  open-source vision-language-action model,'' \emph{arXiv preprint
  arXiv:2406.09246}, 2024.

\bibitem{team2024octo}
O.~M. Team, D.~Ghosh, H.~Walke, K.~Pertsch, K.~Black, O.~Mees, S.~Dasari,
  J.~Hejna, T.~Kreiman, C.~Xu \emph{et~al.}, ``Octo: An open-source generalist
  robot policy,'' \emph{arXiv preprint arXiv:2405.12213}, 2024.

\bibitem{zhang2025ta}
Z.~Zhang, H.~Xu, Z.~Yang, C.~Yue, Z.~Lin, H.-a. Gao, Z.~Wang, and H.~Zhao,
  ``Ta-vla: Elucidating the design space of torque-aware vision-language-action
  models,'' \emph{arXiv preprint arXiv:2509.07962}, 2025.

\bibitem{yu2026forcevla}
J.~Yu, H.~Liu, Q.~Yu, J.~Ren, C.~Hao, H.~Ding, G.~Huang, G.~Huang, Y.~Song,
  P.~Cai \emph{et~al.}, ``Forcevla: Enhancing vla models with a force-aware moe
  for contact-rich manipulation,'' \emph{Advances in Neural Information
  Processing Systems}, vol.~38, pp. 93\,409--93\,439, 2026.

\bibitem{intelligence2025pi}
P.~Intelligence, A.~Amin, R.~Aniceto, A.~Balakrishna, K.~Black, K.~Conley,
  G.~Connors, J.~Darpinian, K.~Dhabalia, J.~DiCarlo \emph{et~al.},
  ``$pi^*_0.6$: a vla that learns from experience,'' \emph{arXiv preprint
  arXiv:2511.14759}, 2025.

\bibitem{li2025simplevla}
H.~Li, Y.~Zuo, J.~Yu, Y.~Zhang, Z.~Yang, K.~Zhang, X.~Zhu, Y.~Zhang, T.~Chen,
  G.~Cui \emph{et~al.}, ``Simplevla-rl: Scaling vla training via reinforcement
  learning,'' \emph{arXiv preprint arXiv:2509.09674}, 2025.

\bibitem{ghasemipour2025self}
S.~K.~S. Ghasemipour, A.~Wahid, J.~Tompson, P.~Sanketi, and I.~Mordatch,
  ``Self-improving embodied foundation models,'' \emph{arXiv preprint
  arXiv:2509.15155}, 2025.

\bibitem{wagenmaker2025steering}
A.~Wagenmaker, M.~Nakamoto, Y.~Zhang, S.~Park, W.~Yagoub, A.~Nagabandi,
  A.~Gupta, and S.~Levine, ``Steering your diffusion policy with latent space
  reinforcement learning,'' \emph{arXiv preprint arXiv:2506.15799}, 2025.

\bibitem{xiao2025self}
W.~Xiao, H.~Lin, A.~Peng, H.~Xue, T.~He, Y.~Xie, F.~Hu, J.~Wu, Z.~Luo, L.~Fan
  \emph{et~al.}, ``Self-improving vision-language-action models with data
  generation via residual rl,'' \emph{arXiv preprint arXiv:2511.00091}, 2025.

\bibitem{mark2024policy}
M.~S. Mark, T.~Gao, G.~G. Sampaio, M.~K. Srirama, A.~Sharma, C.~Finn, and
  A.~Kumar, ``Policy agnostic rl: Offline rl and online rl fine-tuning of any
  class and backbone,'' \emph{arXiv preprint arXiv:2412.06685}, 2024.

\bibitem{wang2602gigabrain}
B.~Wang, B.~Li, C.~Ni, G.~Huang, G.~Zhao, H.~Li, J.~Li, J.~Lv, J.~Liu, L.~Feng
  \emph{et~al.}, ``Gigabrain-0.5 m*: a vla that learns from world model-based
  reinforcement learning, 2026,'' \emph{URL https://arxiv.org/abs/2602.12099},
  2026.

\bibitem{zhu2025wmpo}
F.~Zhu, Z.~Yan, Z.~Hong, Q.~Shou, X.~Ma, and S.~Guo, ``Wmpo: World model-based
  policy optimization for vision-language-action models,'' \emph{arXiv preprint
  arXiv:2511.09515}, 2025.

\bibitem{xiao2025world}
J.~Xiao, Y.~Yang, X.~Chang, R.~Chen, F.~Xiong, M.~Xu, W.-S. Zheng, and
  Q.~Zhang, ``World-env: Leveraging world model as a virtual environment for
  vla post-training,'' \emph{arXiv preprint arXiv:2509.24948}, 2025.

\bibitem{yang2026rise}
J.~Yang, K.~Lin, J.~Li, W.~Zhang, T.~Lin, L.~Wu, Z.~Su, H.~Zhao, Y.-Q. Zhang,
  L.~Chen \emph{et~al.}, ``Rise: Self-improving robot policy with compositional
  world model,'' \emph{arXiv preprint arXiv:2602.11075}, 2026.

\bibitem{xu2026rl}
C.~Xu, J.~T. Springenberg, M.~Equi, A.~Amin, A.~Esmail, S.~Levine, and L.~Ke,
  ``Rl token: Bootstrapping online rl with vision-language-action models,''
  \emph{arXiv preprint arXiv:2604.23073}, 2026.

\bibitem{maes2026leworldmodel}
L.~Maes, Q.~L. Lidec, D.~Scieur, Y.~LeCun, and R.~Balestriero, ``Leworldmodel:
  Stable end-to-end joint-embedding predictive architecture from pixels,''
  \emph{arXiv preprint arXiv:2603.19312}, 2026.

\bibitem{lu2019vilbert}
J.~Lu, D.~Batra, D.~Parikh, and S.~Lee, ``Vilbert: Pretraining task-agnostic
  visiolinguistic representations for vision-and-language tasks,'' in
  \emph{Advances in Neural Information Processing Systems}, 2019.

\bibitem{haarnoja2017reinforcement}
T.~Haarnoja, H.~Tang, P.~Abbeel, and S.~Levine, ``Reinforcement learning with
  deep energy-based policies,'' in \emph{International conference on machine
  learning}.\hskip 1em plus 0.5em minus 0.4em\relax PMLR, 2017, pp. 1352--1361.

\bibitem{haarnoja2018soft}
T.~Haarnoja, A.~Zhou, P.~Abbeel, and S.~Levine, ``Soft actor-critic: Off-policy
  maximum entropy deep reinforcement learning with a stochastic actor,'' in
  \emph{International conference on machine learning}.\hskip 1em plus 0.5em
  minus 0.4em\relax Pmlr, 2018, pp. 1861--1870.

\bibitem{haarnoja2018soft2}
T.~Haarnoja, A.~Zhou, K.~Hartikainen, G.~Tucker, S.~Ha, J.~Tan, V.~Kumar,
  H.~Zhu, A.~Gupta, P.~Abbeel \emph{et~al.}, ``Soft actor-critic algorithms and
  applications,'' \emph{arXiv preprint arXiv:1812.05905}, 2018.

\bibitem{tan2019lxmert}
H.~Tan and M.~Bansal, ``Lxmert: Learning cross-modality encoder representations
  from transformers,'' in \emph{Proceedings of the 2019 Conference on Empirical
  Methods in Natural Language Processing}, 2019.

\bibitem{jaegle2021perceiver}
A.~Jaegle, F.~Gimeno, A.~Brock, O.~Vinyals, A.~Zisserman, and J.~Carreira,
  ``Perceiver: General perception with iterative attention,'' in
  \emph{International Conference on Machine Learning}, 2021.

\end{thebibliography}

\end{document}